\documentclass[11pt, a4paper, logo, copyright]{googledeepmind}

\usepackage[authoryear, sort&compress, round]{natbib}
\usepackage{hyperref}
\usepackage{url}
\usepackage{graphicx}
\usepackage{booktabs}
\usepackage{makecell}
\usepackage{multirow}
\usepackage{amsmath}

\definecolor{gdm_darkgray}{HTML}{9aa0a6}
\definecolor{gdm_gray}{HTML}{3c4043}

\title{Efficient and Robust Drafting with Multi-Token Prediction}

\author{Filippo Galgani}
\author{Ziwei Ji}
\author{Omri Homburger}
\author{Ganesh Jawahar}
\author{Michelle Ramanovich}
\author{Ivan Korotkov}
\author{Luciano Martins}
\author{Utku Evci}
\author{Edouard Yvinec}
\author{Pooja Consul}
\author{Vivek Kumar}
\author{Tal Schuster}
\affil[1]{Google}

\begin{abstract}
Speculative decoding has emerged as a critical technique for accelerating Large Language Model (LLM) inference by decoupling generation into a drafting phase and a parallel verification phase.
A crucial challenge is therefore designing a drafter that is both lightweight and accurate enough to approximate the predictions of the target model. We present a highly optimized Multi-Token Prediction (MTP) architecture that extends existing backbone LLMs to efficiently produce high-quality drafts. Built around a single shared drafting head, our approach features three core architectural optimizations: \textit{target activation injection} to anchor semantic planning, \textit{zero-copy KV-cache cross-attention} to bypass prefill compute and reduce memory overhead, and an \textit{efficient clustered vocabulary projection} to reduce the cost of full-vocabulary decoding. Testing these improvements with the Gemma 4 model family, our G-MTP shows out-of-the-box speedups for both large server-side setups (up to 3.1$\times$) and smaller models targeting memory-constrained on-device environments (up to 2.3$\times$ on H100 and up to 2.2$\times$ on a smartphone GPU). In addition, we present parameter-efficient fine-tuning recipes for G-MTP that keep the drafter aligned with the backbone LLM under domain shift, requiring up to 1.83$\times$ fewer target forward passes than tuning only the backbone.
\end{abstract}

\begin{document}

\maketitle

\section{Introduction}
Deploying Large Language Models (LLMs) at scale — across both server clusters and edge devices — faces a strict bottleneck due to the autoregressive nature of standard transformers. Generating a single token requires a full forward pass, which under-utilizes modern hardware and strains memory bandwidth. Speculative decoding mitigates this using a smaller draft model to predict multiple tokens ahead, which the target model verifies in parallel.
However, standalone draft models must independently process the entire context, introducing additional prefill computations and separate memory overheads. This added friction degrades throughput in server clusters and completely prohibits deployment in memory-constrained edge devices like mobile phones.

To address this, we introduce an optimized Multi-Token Prediction head (G-MTP) that can be fully integrated with any modern LLM, such as Gemma 4. Instead of a separate model, we append a lightweight, shared 4-layer Transformer stack directly to the final layers of the main model. This architecture maximizes token acceptance while minimizing latency and memory overhead through several key innovations:
\begin{itemize}
\item \textbf{Shared Drafting Head:} A single, lightweight transformer stack drafts multiple future tokens autoregressively. Although trained for a 3-step lookahead, the head generalizes to longer draft sequences during inference.
\item \textbf{Target Activation Injection:} The drafter takes the final activation of the target backbone as input, anchoring its planning in deep semantic history, as done by \citet{Li2024}.
\item \textbf{Zero-Copy KV-Cache Cross-Attention:} The drafter cross-attends directly to the target model's KV cache using a constant positional encoding, completely eliminating drafter prefill overhead.
\item \textbf{Clustered Vocabulary Projection:} A two-stage, cluster-based vocabulary projection that drastically reduces the computational matrix-multiplication overhead of projecting draft tokens onto the full vocabulary.
 \item \textbf{Robust Domain Adaptation:} A computationally efficient distillation regime that prevents the drafter from suffering severe distributional decoupling when the verification backbone is fine-tuned for specialized downstream tasks.
\end{itemize}
We trained G-MTP drafters for the full Gemma 4 model family \citep{gemmateam2026gemma4technicalreport} and release the checkpoints.\footnote{\url{https://ai.google.dev/gemma/docs/mtp/mtp}} To support reproducibility, we also release the complete domain adaptation training pipelines described in Section~\ref{sec:sft}.\footnote{\url{https://github.com/google-research/gemma4-mtp-drafter-domain-adaptation}}

\section{Related Work}
Speculative decoding accelerates inference via rapid drafting and parallel verification \citep{Leviathan2023, Chen2023}. Foundational works like blockwise parallel decoding \citep{Stern2018}, Medusa \citep{Cai2024}, and multi-token prediction (MTP) \citep{gloeckle2024better} train multiple independent non-autoregressive heads to achieve self-speculative speedups. More recently, DFlash \citep{chen2026dflash} further optimizes this parallel drafting paradigm to maximize drafting speed. In contrast to parallel heads, EAGLE \citep{Li2024}, alongside recurrent and sequentially dependent drafters \citep{cheng2024recurrentdrafterfastspeculative, ankner2024hydrasequentiallydependentdraftheads}, introduced sequential autoregressive drafting conditioned on prior hidden states and sampled tokens. Similarly, several frontier models adopted a sequential MTP approach, including DeepSeek-V3 \citep{DeepSeek2024}, Gemma 4 \citep{gemmateam2026gemma4technicalreport}, Qwen3 \citep{yang2025qwen3technicalreport}, GLM-4.5 \citep{5team2025glm45agenticreasoningcoding}, and MiMo \citep{xiaomi2025mimounlockingreasoningpotential}. While some prior MTP designs assign separate modules for each lookahead step, our architecture, similar to FastMTP~\citep{cai2025fastmtpacceleratingllminference}, maximizes memory efficiency by reusing a single shared 4-layer block.

A persistent challenge in developing sequential, feature-based draft models is managing the inherent uncertainty of autoregressive generation. As highlighted by EAGLE \citep{Li2024}, continuous feature sequences exhibit significant ambiguity because multiple valid token branches can naturally emerge from a single hidden state. To explicitly resolve this randomness, EAGLE incorporates the token sequence advanced by one time step alongside the feature sequence, anchoring the next-feature prediction with the actual sampled token. Our architecture integrates this insight by dynamically combining the main model's current hidden state with the newly sampled next token's embedding, without requiring complex auxiliary feature-pooling layers.

To bypass the standalone drafter prefill bottleneck, recent architectures loosely couple the drafter to the target's Key-Value (KV) cache~\citep{s2024tandemtransformersinferenceefficient,Du2024} or redesign decoders into minimal cross-attention structures~\citep{Zhong2026}. However, our integrated G-MTP layers directly share the primary KV cache via zero-copy cross-attention, eliminating standalone prefill overhead entirely. Finally, because fine-tuned backbones can cause drafters to undergo severe distributional decoupling \citep{zhou2024distillspec}, we build on distillation-based MTP training and post-training adaptation methods \citep{teoh2026multitokenpredictionright, badarayani2026matchdistributioncomputeposttraining, kirchenbauer2026multitokenpredictionselfdistillation} to demonstrate how G-MTP drafters survive domain shifts via robust, parameter-efficient adaptation strategies.

\section{Methodology and Architecture}
\begin{figure}[htbp]
    \centering
    \includegraphics[width=0.6\linewidth]{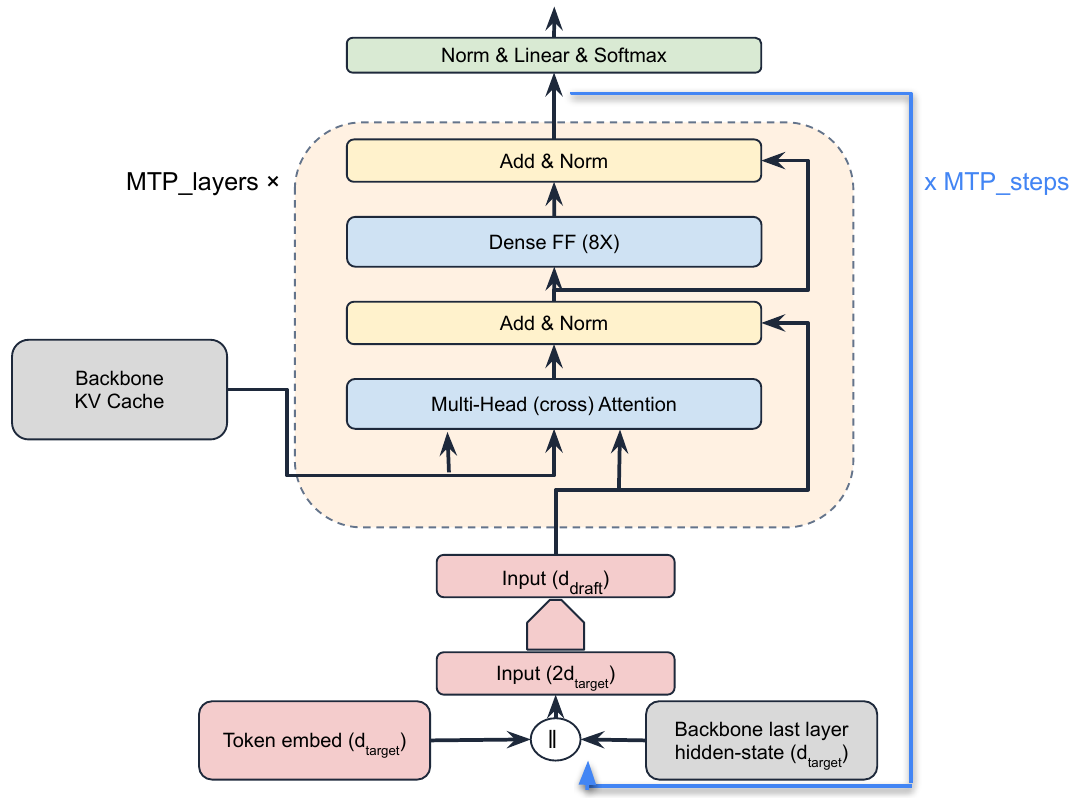}
    \caption{Architecture of the optimized Multi-Token Prediction (G-MTP) module. The input to the drafter is constructed by concatenating the token embedding ($e(x)$) with the backbone hidden state ($h^{(L)}$), both of dimension $d_{\text{target}}$, and reducing to the drafter dimension $d_{\text{draft}}$. The draft tokens are then propagated through the MTP layers, utilizing cross-attention against the backbone's frozen KV Cache to bypass prefill computations before recursively triggering the next forward step. For subsequent drafting steps, the drafter's output state is up-projected ($W_{\text{up}}$), concatenated with the embedding of the sampled draft token, and down-projected ($W_{\text{down}}$) to form the next input (Equation~\ref{eq:zk_recursive}).}
    \label{fig:mtp_architecture}
\end{figure}

\begin{figure}
    \centering
    \includegraphics[width=0.6\linewidth]{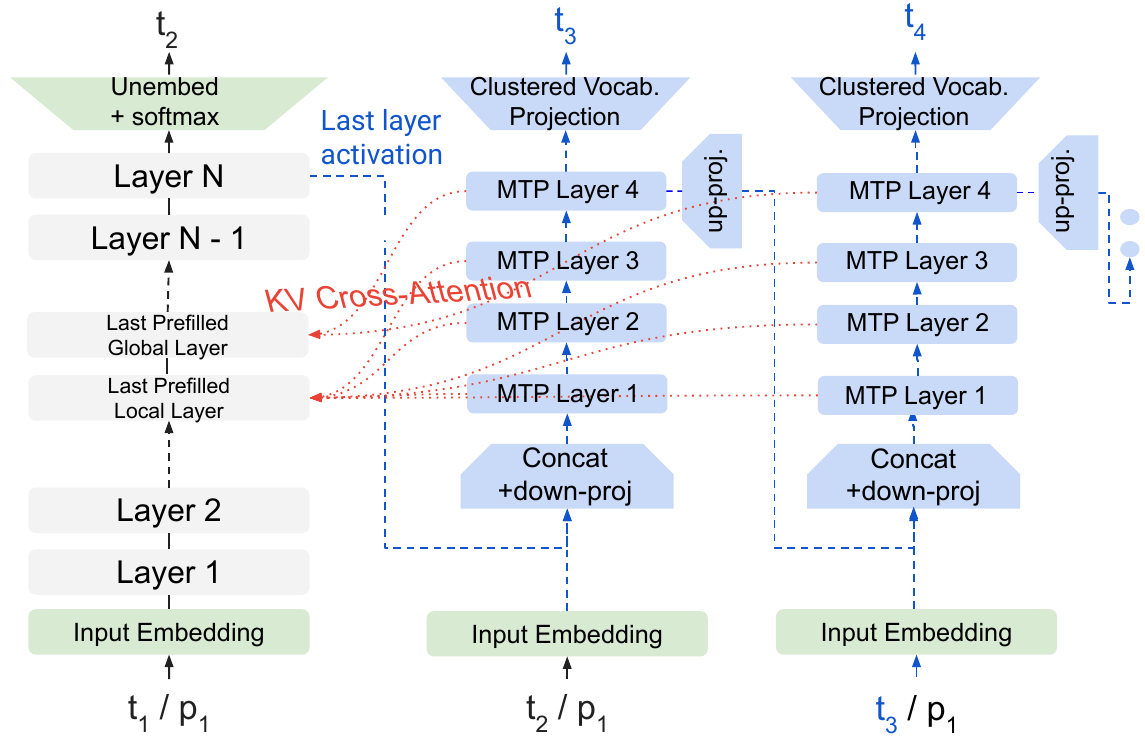}
    \caption{The autoregressive G-MTP drafter (blue blocks) is initialized with activations from the main model (gray blocks) and cross-attends to its KV cache. Across successive drafting steps, the positional encoding ($p_1$) remains constant, as the drafter repeatedly queries a static, pre-computed KV snapshot.}
    \label{fig:mtp}
\end{figure}

At the core of our speculative decoding framework is a shared 4-layer Multi-Token Prediction (G-MTP) Transformer block that outputs candidate tokens autoregressively. To initiate the drafting phase, the target backbone's final layer hidden activations are combined with the current token's embedding from the target backbone's embedding table, and fed into the MTP head. Figure~\ref{fig:mtp_architecture} highlights the construction of this input alongside the cross-attention flow. Rather than building an isolated context representation, the drafter actively queries the primary model's pre-computed Key-Value (KV) cache via \textit{Zero-Copy Cross-Attention} (Figure~\ref{fig:mtp}). This zero-copy memory integration completely bypasses the traditional prefill overhead associated with standalone draft models. Because the same 4-layer stack is recursively applied to predict every subsequent candidate, the architecture naturally accommodates arbitrary lookahead horizons. Following generation, the target backbone evaluates the entire draft trajectory concurrently in a single parallel verification step.

\subsection{Target Activation Injection}
\label{sec:target_activation_injection}
Standalone draft models frequently suffer from lower acceptance rates because they generate predictions based solely on text history, remaining ``blind'' to the rich internal semantics computed by the primary target model. To provide better semantic context, our framework integrates directly with the target model. The drafting process is initialized by extracting the high-dimensional activations from the final processing layers of the target model.

Let $e(x_{t+1}) \in \mathbb{R}^{d_{\text{target}}}$ denote the (scaled) backbone embedding of the newly sampled token and $h_t^{(L)} \in \mathbb{R}^{d_{\text{target}}}$ denote the final-layer hidden state of the target backbone, taken after its final RMSNorm. The initial input to the drafter, $z_1$, is constructed by concatenating these vectors and projecting them down to the smaller internal dimension of the drafter ($d_{\text{draft}}$):
\begin{equation}
    z_1 = W_{\text{down}} \left[ e(x_{t+1}) \parallel h_t^{(L)} \right]
\end{equation}
where $W_{\text{down}} \in \mathbb{R}^{d_{\text{draft}} \times 2d_{\text{target}}}$. For any lookahead step $k \geq 1$, given the projected input vector $z_k \in \mathbb{R}^{d_{\text{draft}}}$, our shared 4-layer drafting block processes $z_k$ by cross-attending directly to the target backbone's prefilled Key-Value caches, producing the next hidden state representation $h_{t+k}^{(\text{draft})} \in \mathbb{R}^{d_{\text{draft}}}$ (the output of the drafter's final RMSNorm):
\begin{equation}
    h_{t+k}^{(\text{draft})} = \text{Transformer}_{\text{draft}}\left(z_k, \, K_{\text{backbone}}, \, V_{\text{backbone}}\right)
\end{equation}
Because the target model has not yet verified the draft sequence, the MTP head acts recurrently for subsequent steps ($k > 1$). Instead of maintaining an auxiliary growing KV-cache for its own predictions, the drafter functions similarly to a Recurrent Neural Network (RNN) over the draft trajectory, explicitly propagating state via its previous hidden activation. Formally, the drafter up-projects its previous hidden state $h_{t+k-1}^{(\text{draft})}$ back to the full target dimension using an up-projection matrix $W_{\text{up}} \in \mathbb{R}^{d_{\text{target}} \times d_{\text{draft}}}$, concatenates it with the shared embedding of its newly predicted draft token $e(\tilde{x}_{t+k})$, and projects it down again to construct the subsequent input $z_k$ (Figure~\ref{fig:mtp_architecture}):
\begin{equation}
    z_k = W_{\text{down}} \left[ e(\tilde{x}_{t+k}) \parallel W_{\text{up}} \, h_{t+k-1}^{(\text{draft})} \right]
    \label{eq:zk_recursive}
\end{equation}
This recursive projection natively maintains the autoregressive causal chain across the drafted sequence without requiring further forward passes through the backbone or storing additional keys and values for the drafted tokens.

\subsection{Zero-Copy KV-Cache Cross-Attention}
A primary inefficiency in standard speculative decoding is the computational cost of prefilling the draft model's KV cache. To completely eliminate this prefill latency, our architecture employs a zero-copy KV cache sharing mechanism, where the G-MTP block cross-attends directly to the already-computed KV cache of the target model.

Mirroring the backbone's interleaved local and global attention with a Local-Local-Local-Global topology, the drafter cross-attends directly to the target's last prefilled local and global layers. Specifically, the drafter's three local layers all cross-attend to the same last prefilled backbone local layer, while its single global layer cross-attends to the last prefilled backbone global layer. When the backbone employs cross-layer KV sharing \citep{brandon2024reducing}, as in E2B and E4B, these ``last prefilled'' caches do not strictly align with the absolute ultimate layer of the backbone. When the drafter cross-attends to the frozen KV cache, it utilizes a constant base positional encoding ($p_1$) for all queries; the forward temporal progression across drafting steps is instead implicitly encoded within the recurrently injected hidden states (Equation~\ref{eq:zk_recursive}) rather than through explicit absolute position IDs. Although multiple sequential drafting layers cross-attend to identically shared static keys and values, they are not redundant; each successive drafting layer dynamically updates its query vectors to extract progressively refined semantic relationships.

This direct cross-attention mechanism achieves a dual benefit: first, it provides the lightweight drafter with deep, full-context semantic history, drastically improving the acceptance rate of drafted tokens on complex tasks. Second, it prevents data duplication in memory. By querying the backbone's KV cache directly, the drafter needs no KV cache of its own, which is especially relevant for memory-constrained on-device deployments.

\subsection{Clustered Vocabulary Projection}
\label{sec:clustered_vocab}
In smaller transformers, like drafters, the final decoding step — multiplying the hidden state by a large embedding matrix to compute logits over the full vocabulary — constitutes a significant latency overhead. In the case of the Gemma 4 architecture, projecting a hidden state to a large vocabulary of $|V| = 262{,}144$ tokens requires $\mathcal{O}(|V| \times d_{\text{draft}})$ operations. For a lightweight drafter, this overhead slows every sequential drafting step, eroding wall-clock speedups gained from speculative decoding.

 To mitigate this latency overhead, we introduce a \textit{clustered vocabulary projection} \citep{zhang2026dynaspeccontextawaredynamicspeculative}. 
 During an offline phase, the drafter's own un-embedding matrix ($|V| \times d_{\text{draft}}$) is clustered into $K=2048$ distinct groups of size 128, using a balanced $K$-means clustering, yielding a representative centroid in $\mathbb{R}^{d_{\text{draft}}}$ for each cluster.
 
 The drafter performs a two-stage sparse routing: it first produces cluster logits ($d_{\text{draft}} \times 2048$ operation, with $d_{\text{draft}}=256$ for E2B and E4B), then computes exact token logits only for the top-32 most probable clusters ($d_{\text{draft}} \times 4096$, i.e., 32 clusters of 128 tokens). This aggressive matrix compression reduces the token-logit computation from $d_{\text{draft}} \times 262{,}144$ to $d_{\text{draft}} \times (2{,}048 + 4{,}096)$ multiply-accumulates per step ($\approx 43\times$ fewer), while maintaining a $>97\%$ token hit rate. On the Edge TPU of a Google Pixel 11, this reduces the drafting time of E2B and E4B by about 20\% (Section~\ref{sec:on_device_npu}).

\section{Experimental Results}
We trained G-MTP drafters for the full Gemma 4 model family \citep{gemmateam2026gemma4technicalreport}. All the drafters have the same architecture as described in this paper, differing only in the model dimension of the drafter and in the vocabulary projection: the Transformer block has model dimension 256 for Gemma 4 E2B and E4B, and 1024 for 26B MoE, 12B and 31B. All have three local and one global attention layers and a feed-forward expansion ratio of 8. The E2B and E4B drafters use the efficient clustered vocabulary projection while the larger models' drafters use a standard full vocabulary projection.

To demonstrate that the theoretical block efficiency gains driven by G-MTP speculative decoding translate to wall-clock acceleration, we evaluate the end-to-end generation throughput of the Gemma 4 model family. Table~\ref{tab:hardware_speedup} presents the autoregressive decoding speed (tokens/s) measured on NVIDIA H100 GPUs (a single H100 GPU for E2B, E4B and 12B, 2$\times$H100 for 26B and 31B) utilizing bfloat16 and the vLLM serving engine (v0.26), with batch size 1, measured on the 880 prompts in the NVIDIA SPEED-Bench quality dataset \citep{abramovich2026speedbenchunifieddiversebenchmark}. See Appendix~\ref{app:gamma_selection} and Table~\ref{tab:gamma_ablation} for a full ablation of $\gamma$.

\begin{table*}[h]
\centering
\caption{Parameter counts and generation speedup across the Gemma 4 family (throughput measured on H100, in tokens per second).}
\label{tab:hardware_speedup}
\resizebox{\textwidth}{!}{%
\begin{tabular}{lccccccc}
\toprule
\textbf{Model} & \textbf{\makecell{Drafter \\ Dimension}} & \textbf{Drafter Params} & \textbf{Optimal $\gamma$} & \textbf{Avg BE} & \textbf{\makecell{Baseline \\ Throughput}} & \textbf{\makecell{Throughput \\ with G-MTP}} & \textbf{Speedup} \\
\midrule
Gemma 4 E2B     & 256  & 76M  & 4  & 2.71 & 278.8 & 550.1 & \textbf{1.97$\times$} \\
Gemma 4 E4B     & 256  & 77M  & 9  & 3.45 & 178.0 & 415.5 & \textbf{2.33$\times$} \\
Gemma 4 12B     & 1024 & 400M & 6  & 3.69 & 93.5  & 249.7 & \textbf{2.67$\times$} \\
Gemma 4 26B MoE & 1024 & 430M & 4  & 3.27 & 248.9 & 334.0 & \textbf{1.34$\times$} \\
Gemma 4 31B     & 1024 & 500M & 8  & 4.26 & 73.2  & 230.1 & \textbf{3.14$\times$} \\
\bottomrule
\end{tabular}%
}
\end{table*}

\begin{table*}[h]
\centering
\caption{Comparison of speedup for three different Gemma 4 31B drafters (2$\times$H100, utilizing FlashAttention 4 kernels, batch size 1) on SPEED-Bench quality dataset.}
\label{tab:gemma_results}
\begin{tabular}{llcrrr}
\toprule
\textbf{Model} & 
\textbf{Mode} & 
\boldmath{$\gamma$} & 
\shortstack{\textbf{Block Eff.}} & 
\shortstack{\textbf{Output TPS}} & 
\textbf{Speedup} \\
\midrule
Gemma 4 31B & Baseline       & N/A & 1.00 & 73.2  & 1.00$\times$ \\
\midrule
Gemma 4 31B & DFlash         & 2   & 2.21 & 142.2 & 1.94$\times$ \\
Gemma 4 31B & DFlash         & 4   & 2.76 & 173.6 & 2.37$\times$ \\
Gemma 4 31B & DFlash         & 8   & 3.15 & 190.5 & 2.60$\times$ \\
\midrule
Gemma 4 31B & EAGLE-3         & 2   & 2.08 & 136.4 & 1.86$\times$ \\
Gemma 4 31B & EAGLE-3         & 4   & 2.52 & 153.8 & 2.10$\times$ \\
Gemma 4 31B & EAGLE-3         & 8   & 2.80 & 151.5 & 2.07$\times$ \\
\midrule
Gemma 4 31B & G-MTP            & 2   & 2.45 & 166.5 & 2.27$\times$ \\
Gemma 4 31B & G-MTP           & 4   & 3.35 & 208.8 & 2.85$\times$ \\
Gemma 4 31B & G-MTP          & 8   & \textbf{4.26} & \textbf{230.1} & \textbf{3.14$\times$} \\
\bottomrule
\end{tabular}%
\end{table*}

While smaller models like the E4B naturally possess high baseline generation speeds (178 tokens/s), deploying the ultra-lightweight G-MTP drafter accelerates it further to 415 tokens/s. This is particularly exciting because the E2B and E4B parameter bounds fall directly within the capacities of modern smartphones, making them strong candidates for on-device speedups. On-device measurements support this: with $\gamma=3$, G-MTP accelerates decoding by 1.69$\times$ for E2B and 2.18$\times$ for E4B on the GPU of a Samsung Galaxy S26 Ultra (Appendix~\ref{app:on_device}), and by 1.92$\times$ and 1.98$\times$ on the Edge TPU of a Google Pixel 11 (Table~\ref{tab:ce_ablation}). More notably, as the verification backbone scales in parameter count (e.g., 12B and 31B models), the relative wall-clock cost of the verification pass amortizes efficiently against the drafter, yielding profound $\sim$2.7$\times$ to 3.1$\times$ accelerations in end-to-end generation throughput, mitigating the memory-bandwidth bottlenecks traditionally associated with autoregressive sampling on large models. 

For the 26B Mixture-of-Experts (MoE) architecture, verification carries a higher overhead per token at small batch sizes due to the potential activation of distinct experts across draft tokens, as evidenced by the high BE translating into a lower speedup. However, when operating at larger batch sizes, these expert activation costs are efficiently amortized across concurrent requests, yielding higher relative throughput speedups that plateau between 1.55$\times$ and 1.71$\times$ for batch sizes of 3 and above (Table~\ref{tab:26b_batch_size_scaling}).

Finally, in Table~\ref{tab:gemma_results}, we compare the measurements obtained for Gemma 4 31B to other speculative drafting frameworks. To ensure a rigorous comparison and eliminate software confounders, all alternative drafters were evaluated under identical conditions to our primary baselines (vLLM serving engine, bfloat16, 2$\times$H100). This comparison is focused on the 31B variant as alternative drafters like EAGLE-3\footnote{\url{https://huggingface.co/RedHatAI/gemma-4-31B-it-speculator.eagle3}} \citep{li2025eagle3scalinginferenceacceleration} and DFlash\footnote{\url{https://huggingface.co/z-lab/gemma-4-31B-it-DFlash}} \citep{chen2026dflash} are readily available, while standard implementations are lacking for the smaller, edge-optimized variants. As shown, G-MTP consistently outperforms both EAGLE-3 and DFlash across all evaluated horizons, achieving superior block efficiency and output TPS. While DFlash can draft faster thanks to its parallel design, it achieves lower block efficiency, leading to lower overall speedups compared to G-MTP.

\paragraph{On-device evaluation on a mobile NPU.}
\label{sec:on_device_npu}
We also evaluate G-MTP on the Edge TPU of a Google Pixel 11, using 4-bit quantized Gemma 4 E2B and E4B, greedy decoding, $\gamma=3$ and at most 600 output tokens, on the 456 single-turn prompts of the SPEED-Bench qualitative set \citep{abramovich2026speedbenchunifieddiversebenchmark}. To measure the on-device effect of the clustered vocabulary projection (Section~\ref{sec:clustered_vocab}), we compare drafters with the full vocabulary projection (no-CE) and with the clustered projection (CE). As shown in Table~\ref{tab:ce_ablation}, G-MTP accelerates decoding by 1.89$\times$ (E2B) and 1.96$\times$ (E4B) without the clustered projection and by 1.92$\times$ and 1.98$\times$ with it. The clustered projection reduces the drafting time to 0.81$\times$ (E2B) and 0.79$\times$ (E4B) of the full-projection drafting time, but it also lowers the median acceptance rate (from 0.49 to 0.48 for E2B and from 0.52 to 0.50 for E4B), which offsets most of this saving at $\gamma=3$: the net end-to-end gain is 1.01$\times$ for both models.

\begin{table*}[h]
\centering
\caption{Full (no-CE) vs.\ clustered (CE) vocabulary projection on the Edge TPU of a Google Pixel 11 (SPEED-Bench qualitative, 456 prompts, $\gamma=3$, greedy decoding). Values are medians over prompts; speedups are ratios of median decoding throughput. CE vs.\ no-CE columns are ratios of decoding speedup and of total drafting time.}
\label{tab:ce_ablation}
\begin{tabular}{lcccc}
\toprule
\textbf{Model} & \textbf{\makecell{no-CE Speedup \\ vs.\ Baseline}} & \textbf{\makecell{CE Speedup \\ vs.\ Baseline}} & \textbf{\makecell{CE vs.\ no-CE \\ Speedup}} & \textbf{\makecell{CE vs.\ no-CE \\ Drafter Time}} \\
\midrule
Gemma 4 E2B & 1.89$\times$ & 1.92$\times$ & 1.01$\times$ & 0.81$\times$ \\
Gemma 4 E4B & 1.96$\times$ & 1.98$\times$ & 1.01$\times$ & 0.79$\times$ \\
\bottomrule
\end{tabular}%
\end{table*}

\section{Speculative Decoding Adaptation via Multi-Token Prediction Distillation}
\label{sec:sft}

To investigate the dynamics of speculative decoding during domain adaptation, we conduct experiments using the Gemma 4 E4B model as our target verification backbone ($\mathcal{M}_{\text{target}}$). For speculative drafting, we utilize the open-sourced G-MTP drafter ($\mathcal{M}_{\text{drafter}}$).

To test our fine-tuning strategies across diverse structural shifts, we evaluate on two different output domains: (1)~\textbf{\texttt{orkg/SciQA}} \citep{SciQA2023}: a semantic parsing dataset that maps natural language questions to complex SPARQL database queries over the Open Research Knowledge Graph; and (2)~\textbf{\texttt{GSM8K}} \citep{cobbe2021gsm8k}: a grade-school mathematical reasoning benchmark.

We formalize and compare three distinct domain adaptation regimes for speculative decoding systems, using the original pre-trained target and drafter without any fine-tuning as the reference (\textit{No SFT reference}, row 1 of Table~\ref{tab:spec_eval}). The regimes below are numbered according to their rows in Table~\ref{tab:spec_eval}:

\begin{enumerate}
\setcounter{enumi}{1}
    \item \textbf{Target Fine-Tuned Only (SFT)}: Only the target backbone $\mathcal{M}_{\text{target}}$ is fine-tuned via standard autoregressive cross-entropy on the downstream dataset. The speculative drafter $\mathcal{M}_{\text{drafter}}$ remains frozen in its pre-trained state.
    \item \textbf{Joint Fine-Tuning (SFT + MTP)}: Both $\mathcal{M}_{\text{target}}$ and $\mathcal{M}_{\text{drafter}}$ are fine-tuned simultaneously from base checkpoints using a composite objective that combines SFT with drafter distillation.
    \item \textbf{Two-Stage Distillation}: The domain-adapted target model from regime 2 is kept \textbf{frozen}. The speculative drafter $\mathcal{M}_{\text{drafter}}$ is then trained via probability distillation to align with the target model's updated representations.
\end{enumerate}

During training ($\gamma_{\text{train}}=3$ lookahead), our objective combines the backbone's cross-entropy with an MTP head loss that interpolates between next-token prediction and probability distillation (Total Variation Distance); in all reported experiments we use pure distillation ($w_{\text{distill}}=1.0$, Appendix~\ref{app:domain_fine_tuning}). For inference evaluation ($\gamma=7$ lookahead), we log every verification step across a test set of $N=500$ samples to record NTP Accuracy, Acceptance Rate, Block Efficiency ($\beta$), and total Target Steps. All model updates are performed via Low-Rank Adaptation (LoRA) \citep{hu2022lowrank}, with full details provided in Appendix~\ref{app:domain_fine_tuning}.

Table~\ref{tab:spec_eval} details the performance of all domain adaptation regimes evaluated across both domains.

\begin{table*}[h]
\centering
\caption{Speculative Decoding Domain Adaptation Performance (Gemma 4 E4B). Fine-tuning the model without updating the MTP (row 2) can lead to regression in Block Efficiency ($\beta$). Joint fine-tuning (row 3) matches or exceeds the $\beta$ of the no SFT reference (row 1), and two-stage distillation (row 4) recovers about 96\% of it.}
\label{tab:spec_eval}
\resizebox{\textwidth}{!}{%
\begin{tabular}{c lcccc}
\toprule
\textbf{Dataset} & \textbf{Configuration Regime} & \textbf{NTP Accuracy} & \textbf{Acceptance Rate} & \textbf{$\beta$} & \textbf{Target Steps} \\
\midrule
\multirow{4}{*}{\textbf{SciQA}} 
& 1. No SFT reference & 68.40\% & 22.36\% & 2.51 & 12,334 \\
\cmidrule{2-6}
& 2. Target Fine-Tuned Only (SFT) & 99.23\% & 8.96\% & 1.60 & 19,423 \\
& 3. Joint Fine-Tuning (SFT + MTP) & \textbf{99.37\%} & \textbf{28.64\%} & \textbf{2.96} & \textbf{10,595} \\
& 4. Two-Stage Distillation & 99.23\% & 26.48\% & 2.80 & 11,155 \\
\midrule
\multirow{4}{*}{\textbf{GSM8K}} 
& 1. No SFT reference & 74.83\% & 40.17\% & 3.73 & 8,414 \\
\cmidrule{2-6}
& 2. Target Fine-Tuned Only (SFT) & 88.01\% & 28.67\% & 2.95 & 10,435 \\
& 3. Joint Fine-Tuning (SFT + MTP) & 88.00\% & \textbf{40.66\%} & \textbf{3.77} & \textbf{8,143} \\
& 4. Two-Stage Distillation & 88.01\% & 37.90\% & 3.58 & 8,591 \\
\bottomrule
\end{tabular}%

}
\end{table*}

When the target backbone is fine-tuned alone (row 2), Next-Token Prediction (NTP) accuracy jumps significantly on both domains (+30.83 percentage points on SciQA, +13.18 on GSM8K). However, the frozen pre-trained drafter suffers severe distributional decoupling. Consequently, Block Efficiency drops from 2.51 to 1.60 on SciQA (-36.3\%), and from 3.73 to 2.95 on GSM8K (-20.9\%). 

Both Joint Fine-Tuning (row 3) and Two-Stage Distillation (row 4) successfully realign the drafter with the backbone. As shown in Table~\ref{tab:spec_eval}, both adaptation strategies recover most or all of the block efficiency of the no SFT reference (row 1). Given the small difference between the two drafter adaptation methods, we conclude that both domain adaptation frameworks are highly effective mechanisms for maintaining speculative drafting efficiency under target distribution shift.
Despite the drafter being trained to predict only $\gamma_{\text{train}}=3$ future steps, our architecture empowers it to recursively extrapolate its predictions beyond the training horizon (e.g., to an extended $\gamma=7$) and successfully reach high acceptance rates.

\section{Discussion and Conclusion}

We introduced an optimized Multi-Token Prediction (G-MTP) architecture for efficient speculative drafting directly integrated into the main model. By leveraging target activation injection and zero-copy KV-cache cross-attention, our approach bypasses the standalone drafter prefill latency and maintains high acceptance rates. The clustered vocabulary projection further reduces the per-step drafting cost, enabling competitive and robust wall-clock generation speedups across the Gemma 4 family (up to $3.14\times$). Furthermore, our domain adaptation analysis highlights that simple distillation effectively realigns G-MTP drafters under distribution shift, preserving high block efficiencies where fine-tuning only the backbone degrades them. To ensure reproducibility and facilitate ongoing research, the code and distillation recipes for these domain adaptation experiments are made publicly available.\footnote{\url{https://github.com/google-research/gemma4-mtp-drafter-domain-adaptation}} 

\paragraph{Limitations and Future Work} While our architectural innovations—specifically the KV-cache sharing and clustered vocabulary projection—were designed to bypass on-device constraints, our comparisons across the full model family were conducted on server GPUs (H100) to standardize our evaluation baselines. Our on-device measurements cover only E2B and E4B on two devices (a smartphone GPU without the clustered projection, and a mobile NPU with and without it). A natural direction for future work is a broader on-device study across devices, backends and batch sizes that also quantifies the memory and battery benefits of each component. Finally, this paper focuses on the drafter architecture and its adaptation to fine-tuned backbones; the pre-training recipe of the released G-MTP drafters is out of scope. All domain adaptation experiments in Section~\ref{sec:sft} start from the publicly released drafter checkpoints and can therefore be reproduced without it.

\section*{Acknowledgments}
We wish to thank Eden Cohen, Niv Giladi, Ethan Baron, Alon Levkovitch, Omer Vandsburger, Eliya Nachmani, Nadav Bar, Pauline Sho, Joe Zou, Ryan Mullins, Sindhu Raghuram, Mayank Chaturvedi, and Mohammadreza Heydary for their help, support, and feedback on the research.

\bibliography{main}

\appendix
\section{Appendix}
\subsection{Gamma Selection and Batch Size Scaling}
\label{app:gamma_selection}

To determine the optimal lookahead horizon ($\gamma$) for the main generation speedup comparisons presented in Table~\ref{tab:hardware_speedup}, we empirically evaluated the Block Efficiency (BE), resulting speculative throughput (tokens/s), and overall wall-clock speedup across different values of $\gamma$. Table~\ref{tab:gamma_ablation} details these measurements for each model in the Gemma 4 family. 

\begin{figure}[h]
    \centering
    \includegraphics[width=\textwidth]{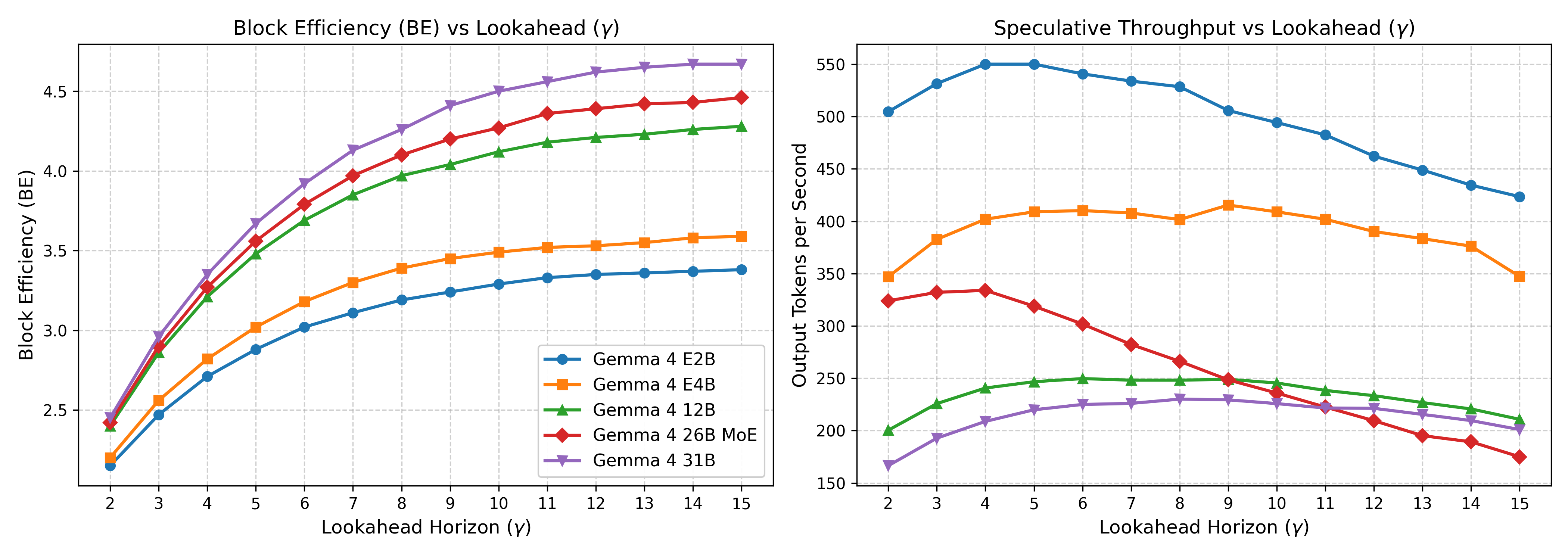}
    \caption{Empirical scaling of theoretical Block Efficiency (BE) and speculative wall-clock throughput across varying lookahead horizons ($\gamma$). While longer draft sequences monotonically increase token acceptance, throughput degrades for some models (e.g., 26B MoE) as drafting and verification costs exceed generation gains.}
    \label{fig:gamma_ablation_plot}
\end{figure}

As $\gamma$ increases, the theoretical block efficiency monotonically rises due to longer draft sequences; however, wall-clock throughput eventually degrades as the penalty of drafting and verifying longer, lower-probability draft tokens outweighs the accepted generation lengths. The optimal $\gamma$ varies heavily depending on the model's structural capacity to parallelize verification (Figure~\ref{fig:gamma_ablation_plot}).

While optimal $\gamma$ dictates the length of the draft sequence, the batch size heavily influences the computational amortization of the verification phase, particularly for Mixture-of-Experts (MoE) architectures running highly concurrent workloads. As detailed in Table~\ref{tab:26b_batch_size_scaling}, we evaluate throughput scaling for the Gemma 4 26B MoE at $\gamma=4$ across batch sizes 1 through 15. At low batch settings, the relative verification overhead of routing discrete draft tokens through parallel experts yields moderate speedups of 1.34$\times$ (batch size 1) and 1.41$\times$ (batch size 2). As the batch size increases, the cost of expert activation is amortized across sequences, and the speedup plateaus between 1.55$\times$ and 1.71$\times$ for batch sizes of 3 and above.

\begin{table*}[h]
\centering
\caption{Throughput across lookahead horizons $\gamma$ (Figure~\ref{fig:gamma_ablation_plot} extends to $\gamma=15$).}
\label{tab:gamma_ablation}
\resizebox{\textwidth}{!}{%
\begin{tabular}{lccccc}
\toprule
\textbf{Model} & \textbf{$\gamma$} & \textbf{Avg BE} & \textbf{Baseline (tokens/s)} & \textbf{G-MTP Speculative (tokens/s)} & \textbf{Throughput Speedup} \\
\midrule
\multirow{11}{*}{Gemma 4 E2B}
 & 2  & 2.15 & 278.8 & 504.7 & 1.81$\times$ \\
 & 3  & 2.47 & 278.8 & 531.5 & 1.91$\times$ \\
 & 4  & 2.71 & 278.8 & \textbf{550.1} & \textbf{1.97$\times$} \\
 & 5  & 2.88 & 278.8 & \textbf{550.1} & \textbf{1.97$\times$} \\
 & 6  & 3.02 & 278.8 & 540.8 & 1.94$\times$ \\
 & 7  & 3.11 & 278.8 & 533.9 & 1.92$\times$ \\
 & 8  & 3.19 & 278.8 & 528.5 & 1.90$\times$ \\
 & 9  & 3.24 & 278.8 & 505.8 & 1.81$\times$ \\
 & 10  & 3.29 & 278.8 & 494.3 & 1.77$\times$ \\
 & 11  & 3.33 & 278.8 & 482.5 & 1.73$\times$ \\
 & 12  & 3.35 & 278.8 & 462.2 & 1.66$\times$ \\
 \midrule
\multirow{11}{*}{Gemma 4 E4B}
 & 2  & 2.20 & 178.0 & 347.0 & 1.95$\times$ \\
 & 3  & 2.56 & 178.0 & 382.7 & 2.15$\times$ \\
 & 4  & 2.82 & 178.0 & 402.0 & 2.26$\times$ \\
 & 5  & 3.02 & 178.0 & 409.0 & 2.30$\times$ \\
 & 6  & 3.18 & 178.0 & 410.2 & 2.30$\times$ \\
 & 7  & 3.30 & 178.0 & 407.9 & 2.29$\times$ \\
 & 8  & 3.39 & 178.0 & 401.6 & 2.26$\times$ \\
 & 9  & 3.45 & 178.0 & \textbf{415.5} & \textbf{2.33$\times$} \\
 & 10  & 3.49 & 178.0 & 409.0 & 2.30$\times$ \\
 & 11  & 3.52 & 178.0 & 401.9 & 2.26$\times$ \\
 & 12  & 3.53 & 178.0 & 390.2 & 2.19$\times$ \\
\midrule
\multirow{11}{*}{Gemma 4 12B}
 & 2  & 2.40 & 93.5 & 200.4 & 2.14$\times$ \\
 & 3  & 2.86 & 93.5 & 225.8 & 2.41$\times$ \\
 & 4  & 3.21 & 93.5 & 240.7 & 2.57$\times$ \\
 & 5  & 3.48 & 93.5 & 246.7 & 2.64$\times$ \\
 & 6  & 3.69 & 93.5 & \textbf{249.7} & \textbf{2.67$\times$} \\
 & 7  & 3.85 & 93.5 & 248.2 & 2.65$\times$ \\
 & 8  & 3.97 & 93.5 & 248.2 & 2.65$\times$ \\
 & 9  & 4.04 & 93.5 & 249.0 & 2.66$\times$ \\
 & 10  & 4.12 & 93.5 & 245.5 & 2.63$\times$ \\
 & 11  & 4.18 & 93.5 & 238.4 & 2.55$\times$ \\
 & 12  & 4.21 & 93.5 & 233.4 & 2.50$\times$ \\
\midrule
\multirow{11}{*}{Gemma 4 26B MoE}
 & 2  & 2.42 & 248.9 & 324.0 & 1.30$\times$ \\
 & 3  & 2.90 & 248.9 & 332.1 & 1.33$\times$ \\
 & 4  & 3.27 & 248.9 & \textbf{334.0} & \textbf{1.34$\times$} \\
 & 5  & 3.56 & 248.9 & 318.9 & 1.28$\times$ \\
 & 6  & 3.79 & 248.9 & 301.8 & 1.21$\times$ \\
 & 7  & 3.97 & 248.9 & 282.3 & 1.13$\times$ \\
 & 8  & 4.10 & 248.9 & 266.2 & 1.07$\times$ \\
 & 9  & 4.20 & 248.9 & 248.5 & 1.00$\times$ \\
 & 10  & 4.27 & 248.9 & 235.9 & 0.95$\times$ \\
 & 11  & 4.36 & 248.9 & 222.5 & 0.89$\times$ \\
 & 12  & 4.39 & 248.9 & 209.5 & 0.84$\times$ \\
\midrule
\multirow{11}{*}{Gemma 4 31B}
 & 2  & 2.45 & 73.2 & 166.5 & 2.27$\times$ \\
 & 3  & 2.96 & 73.2 & 192.8 & 2.63$\times$ \\
 & 4  & 3.35 & 73.2 & 208.8 & 2.85$\times$ \\
 & 5  & 3.67 & 73.2 & 219.9 & 3.00$\times$ \\
 & 6  & 3.92 & 73.2 & 225.0 & 3.07$\times$ \\
 & 7  & 4.13 & 73.2 & 226.0 & 3.09$\times$ \\
 & 8  & 4.26 & 73.2 & \textbf{230.1} & \textbf{3.14$\times$} \\
 & 9  & 4.41 & 73.2 & 229.4 & 3.13$\times$ \\
 & 10  & 4.50 & 73.2 & 225.8 & 3.08$\times$ \\
 & 11  & 4.56 & 73.2 & 221.6 & 3.03$\times$ \\
 & 12  & 4.62 & 73.2 & 221.3 & 3.02$\times$ \\
\bottomrule
\end{tabular}%
}
\end{table*}

\begin{table*}[h]
\centering
\caption{Batch size scaling performance: Gemma 4 26B MoE ($\gamma=4$, measured on 2$\times$H100)}
\label{tab:26b_batch_size_scaling}
\resizebox{\textwidth}{!}{%
\begin{tabular}{lcccccc}
\toprule
\textbf{Model} & \textbf{Batch Size} & \textbf{$\gamma$} & \textbf{Avg BE} & \textbf{Baseline Throughput} & \textbf{Throughput with G-MTP} & \textbf{Speedup} \\
\midrule
Gemma 4 26B MoE & 1  & 4 & 3.27 & 248.9  & 334.0  & 1.34$\times$ \\
Gemma 4 26B MoE & 2  & 4 & 3.27 & 442.8  & 625.9  & 1.41$\times$ \\
Gemma 4 26B MoE & 3  & 4 & 3.27 & 590.0  & 923.7  & 1.57$\times$ \\
Gemma 4 26B MoE & 4  & 4 & 3.27 & 747.9  & 1180.1 & 1.58$\times$ \\
Gemma 4 26B MoE & 5  & 4 & 3.27 & 849.9  & 1409.0 & 1.66$\times$ \\
Gemma 4 26B MoE & 6  & 4 & 3.28 & 982.2  & 1524.6 & 1.55$\times$ \\
Gemma 4 26B MoE & 7  & 4 & 3.27 & 1053.3 & 1797.4 & \textbf{1.71$\times$} \\
Gemma 4 26B MoE & 8  & 4 & 3.28 & 1183.0 & 1965.7 & 1.66$\times$ \\
Gemma 4 26B MoE & 9  & 4 & 3.27 & 1244.2 & 2023.4 & 1.63$\times$ \\
Gemma 4 26B MoE & 10 & 4 & 3.27 & 1366.1 & 2259.3 & 1.65$\times$ \\
Gemma 4 26B MoE & 11 & 4 & 3.27 & 1419.7 & 2297.0 & 1.62$\times$ \\
Gemma 4 26B MoE & 12 & 4 & 3.27 & 1527.2 & 2518.0 & 1.65$\times$ \\
Gemma 4 26B MoE & 13 & 4 & 3.27 & 1571.0 & 2601.7 & 1.66$\times$ \\
Gemma 4 26B MoE & 14 & 4 & 3.27 & 1635.3 & 2622.1 & 1.60$\times$ \\
Gemma 4 26B MoE & 15 & 4 & 3.27 & 1720.8 & 2818.5 & 1.64$\times$ \\
\bottomrule
\end{tabular}%
}
\end{table*}

\subsection{On-Device Measurements}
\label{app:on_device}

Table~\ref{tab:on_device} complements the server-side evaluation with on-device measurements of the G-MTP drafters for Gemma 4 E2B and E4B, deployed with the LiteRT-LM runtime on the GPU of a Samsung Galaxy S26 Ultra. These runs use the quantized LiteRT-LM releases of the models, greedy decoding, a lookahead of $\gamma=3$ (matching the training lookahead) and the full vocabulary projection, i.e., the clustered vocabulary projection is not enabled. The benchmark contains 76 short prompts (median 29 input tokens) from four task types (8 free-form generation, 30 tone rewriting, 10 summarization and 28 code generation prompts), with generation capped at 512 tokens; every prompt is run with and without G-MTP.

G-MTP accelerates decoding by 1.69$\times$ for E2B and 2.18$\times$ for E4B. These speedups are in the same range as those measured on H100 at the same lookahead (1.91$\times$ and 2.15$\times$; Table~\ref{tab:gamma_ablation}), although the prompt sets differ, and, as on H100, E2B obtains a smaller speedup than E4B. For both models, free-form generation has the lowest acceptance rate and the smallest speedup.

\begin{table*}[h]
\centering
\caption{On-device decoding throughput of Gemma 4 E2B and E4B with LiteRT-LM on the GPU of a Samsung Galaxy S26 Ultra ($\gamma=3$, greedy decoding, full vocabulary projection). Acceptance rate and throughput values are medians over the 76 prompts; the acceptance rate is the draft acceptance rate reported by the runtime. Speedups are ratios of median throughput with and without G-MTP, over all prompts and per task type.}
\label{tab:on_device}
\resizebox{\textwidth}{!}{%
\begin{tabular}{lcccccccc}
\toprule
 & \multicolumn{4}{c}{\textbf{All prompts}} & \multicolumn{4}{c}{\textbf{Speedup by task type}} \\
\cmidrule(lr){2-5} \cmidrule(lr){6-9}
\textbf{Model} & \textbf{\makecell{Acceptance \\ Rate}} & \textbf{\makecell{Baseline \\ (tokens/s)}} & \textbf{\makecell{G-MTP \\ (tokens/s)}} & \textbf{Speedup} & \textbf{Free-form} & \textbf{Rewrite} & \textbf{Summarize} & \textbf{Code} \\
\midrule
Gemma 4 E2B & 0.54 & 50.5 & 85.4 & \textbf{1.69$\times$} & 1.36$\times$ & 1.73$\times$ & 1.78$\times$ & 1.67$\times$ \\
Gemma 4 E4B & 0.59 & 21.6 & 47.1 & \textbf{2.18$\times$} & 1.71$\times$ & 2.22$\times$ & 2.07$\times$ & 2.25$\times$ \\
\bottomrule
\end{tabular}%
}
\end{table*}

\clearpage
\subsection{Details on Domain Fine-Tuning}
\label{app:domain_fine_tuning}
\label{app:peft_details}

\subsubsection{Parameter-Efficient Fine-Tuning}
To ensure computational efficiency and prevent catastrophic forgetting during domain specialization, all model updates are performed via Low-Rank Adaptation (\textbf{LoRA}). We inject rank-$r=8$ adapters ($\alpha = 16$, $\text{dropout} = 0.05$, no bias) into:
\begin{itemize}
    \item \textbf{Target Backbone ($\mathcal{M}_{\text{target}}$)}: Attention projections (\texttt{q\_proj}, \texttt{k\_proj}, \texttt{v\_proj}, \texttt{o\_proj}). This introduces \textbf{5,914,624 trainable parameters} ($\sim$0.07\% of the target backbone).
    \item \textbf{Speculative Drafter ($\mathcal{M}_{\text{drafter}}$)}: Attention and MLP projections (\texttt{q\_proj}, \texttt{o\_proj}, \texttt{gate\_proj}, \texttt{up\_proj}, \texttt{down\_proj}), as well as internal state mapping layers (\texttt{pre\_projection} corresponding to $W_{\text{down}}$, and \texttt{post\_projection} corresponding to $W_{\text{up}}$ from Section~\ref{sec:target_activation_injection}). This introduces \textbf{385,024 trainable parameters} ($\sim$0.5\% of the drafter head). During fine-tuning and evaluation, the drafter keeps its clustered vocabulary projection with the original centroids: logits are computed only for tokens in the top-32 clusters, and all other tokens receive a constant minimum logit; the TVD loss is computed over the full $262$K vocabulary of the resulting distribution. Since the output projection is not updated, the offline clusters remain valid after fine-tuning.
\end{itemize}

Under the Joint Fine-Tuning regime, the total overall system capacity mapped to trainable adapters is bounded to \textbf{6,299,648 parameters}. All models are trained for 3 epochs using AdamW ($\beta_1 = 0.9, \beta_2 = 0.999$, weight decay $= 0.01$) with a peak learning rate of $\eta = 2 \times 10^{-4}$ for SciQA and $\eta = 1 \times 10^{-4}$ for GSM8K (using \texttt{bfloat16} mixed precision, effective batch size $= 16$, linear warmup/cosine decay, gradient clipping at $\|g\|_2 = 0.3$).

\subsubsection{Speculative G-MTP Objective \& Domain Adaptation}
For a training lookahead horizon of $\gamma_{\text{train}}$ steps ($\gamma_{\text{train}}=3$ in our experiments), the speculative drafter takes the target model's hidden representation $h_t^{(L)}$ at step $t$ and predicts candidate distributions across $t+2, \dots, t+\gamma_{\text{train}}+1$. The unified training objective is:
\begin{equation}
    \mathcal{L}_{\text{total}} = \mathcal{L}_{\text{backbone\_LM}} + \lambda_{\text{MTP}} \sum_{k=1}^{\gamma_{\text{train}}} \mathcal{L}_{\text{head\_}k}
\end{equation}
where $\mathcal{L}_{\text{backbone\_LM}}$ is the standard next-token cross-entropy and $\lambda_{\text{MTP}} = 0.4$. For each lookahead step $k$ (i.e., the $k$-th recursive step of the single shared drafter), the head loss $\mathcal{L}_{\text{head\_}k}$ interpolates between ground-truth cross-entropy ($\mathcal{L}_{\text{NTP}}$) and Probability Distillation ($\mathcal{L}_{\text{TVD}}$):
\begin{equation}
    \mathcal{L}_{\text{head\_}k} = (1 - w_{\text{distill}}) \mathcal{L}_{\text{NTP\_}k} + w_{\text{distill}} \mathcal{L}_{\text{TVD\_}k}
\end{equation}
Here, $\mathcal{L}_{\text{NTP\_}k} = -\log P_{\text{drafter\_}k}(x_{t+k+1} \mid h_t^{(L)}, x_{\le t}, x_{t+1:t+k})$ computes the standard autoregressive cross-entropy. Note that during training the drafter is teacher-forced: the tokens $x_{t+1:t+k}$ fed back as drafter inputs are the ground-truth tokens, whereas at inference time the drafter consumes its own predictions $\tilde{x}_{t+k}$ (Equation~\ref{eq:zk_recursive}). For all speculative adaptation experiments reported, we set $w_{\text{distill}} = 1.0$, dedicating the drafter entirely to teacher mimicking. The TVD term minimizes the distance against the target's updated distribution:
\begin{equation}
    \mathcal{L}_{\text{TVD\_}k} = \frac{1}{2} \sum_{v \in \mathcal{V}} \left| P_{\text{drafter\_}k}(v \mid h_t^{(L)}, x_{\le t}, x_{t+1:t+k}) - P_{\text{target}}(v \mid x_{< t+k+1}) \right|
\end{equation}
To strictly isolate the G-MTP updates from warping the target backbone's representations during Joint Fine-Tuning, we apply a stop-gradient over the connection spanning the backbone's initial hidden state $h_t^{(L)}$ and the shared KV cache.

\subsubsection{Evaluation Metrics}
Given accepted tokens ($M_s$) and proposed tokens ($P_s$) during forward pass $s \in \{1, \dots, S_{\text{total}}\}$:
\begin{itemize}
    \item \textbf{NTP Accuracy ($\mathcal{A}_{\text{NTP}}$)}: The teacher-forced next-token prediction accuracy, defined as the frequency with which the target model's top-1 prediction matches the ground-truth token.
    \item \textbf{Acceptance Rate ($\alpha$)}: $\alpha = \frac{\sum M_s}{\sum P_s}$.
    \item \textbf{Block Efficiency ($\beta$)}: $\beta = 1 + \frac{\sum M_s}{S_{\text{total}}}$.
    \item \textbf{Target Steps}: Total target forward passes $S_{\text{total}}$ required to decode the entire test set (lower indicates faster generation).
\end{itemize}

\end{document}